\documentclass[conference]{IEEEtran}

\usepackage{graphicx}
\usepackage{amsmath,amssymb}
\usepackage{siunitx}
\usepackage{upgreek}
\usepackage[noend]{algpseudocode}
\usepackage{algorithm}
\usepackage{multirow}
\usepackage{url}
\usepackage{balance}
\let\labelindent\relax
\usepackage{enumitem}
\usepackage{xcolor}
\usepackage{soul}
\usepackage{hyperref}
\usepackage{caption}
\usepackage{subcaption}

\ifCLASSINFOpdf
\else
\fi

\begin{document}
%\title{Developing Low-Cost Multispectral Imagers using Inter-Band Redundancy Analysis and Greedy Spectral Selection in Hyperspectral Imaging}
% \title{ Management Zones Clustering Using Neural Network-Generated Nitrogen Response Curves}
\title{Counterfactual Analysis of Neural Networks Used to Create Fertilizer Management Zones}
%Simultaneous Prediction Interval and Target Estimation for Deep Learning}

% Developing Low-Cost Multispectral Imagers for Image Classification with Compact Convolutional Neural Networks

% Developing Low-Cost Multispectral Imagers Using Feature 

\author{\IEEEauthorblockN{Giorgio Morales and John Sheppard}
\IEEEauthorblockA{Gianforte School of Computing \\
Montana State University\\
Bozeman, MT 59717}
}

% Anonymous authors
% \author{\IEEEauthorblockN{Anonymous Authors}}

\maketitle

%===========================================================
\begin{abstract}

In Precision Agriculture, the utilization of management zones (MZs) that take into account within-field variability facilitates effective fertilizer management. 
This approach enables the optimization of nitrogen (N) rates to maximize crop yield production and enhance agronomic use efficiency.
However, existing works often neglect the consideration of responsivity to fertilizer as a factor influencing MZ determination. 
In response to this gap, we present a MZ clustering method based on fertilizer responsivity.
We build upon the statement that the responsivity of a given site to the fertilizer rate is described by the shape of its corresponding N fertilizer-yield response (N-response) curve.
Thus, we generate N-response curves for all sites within the field using a convolutional neural network (CNN).
The shape of the approximated N-response curves is then characterized using functional principal component analysis.
Subsequently, a counterfactual explanation (CFE) method is applied to discern the impact of various variables on MZ membership. 
The genetic algorithm-based CFE solves a multi-objective optimization problem and aims to identify the minimum combination of features needed to alter a site's cluster assignment.
Results from two yield prediction datasets indicate that the features with the greatest influence on MZ membership are associated with terrain characteristics that either facilitate or impede fertilizer runoff, such as terrain slope or topographic aspect.
\footnote{This paper is a preprint (accepted to appear in the International Joint Conference on Neural Networks 2024). IEEE copyright notice. 2024 IEEE. Personal use of this material is permitted. Permission from IEEE must be obtained for all other uses, in any current or future media, including reprinting/republishing this material for advertising or promotional purposes, creating new collective works, for resale or redistribution to servers or lists, or reuse of any copyrighted.}
\end{abstract}

\begin{IEEEkeywords}
Neural network response curves, management zones, counterfactual explanations, explainable machine learning%, precision agriculture.
\end{IEEEkeywords}

%===========================================================
\section{Introduction} \label{sec:intro}

In precision agriculture (PA), management zones (MZs) are distinct sub-regions in a field with similar yield-influencing factors~\cite{Khosla513}.
Different MZs account for the variability of factors within the field (e.g., soil composition) and, thus, vary in their requirements for specific treatments.
These zones are areas with relative homogeneity where specific crop management practices are implemented, aiming to optimize crop productivity and reduce the environmental impact by reducing the overall fertilizer applied~\cite{managementNrate,managementNrate2}.

Several methods have been proposed for the delineation of MZs.
Some of them rely on historical yield data solely~\cite{yieldpatterns,yieldpatterns2} while others use information extracted from remote sensing data exclusively~\cite{automaticMZ,MZmultilayer}.
Alternatively, certain approaches employ a combination of covariate factors, encompassing various soil properties, environmental factors, and topographic information~\cite{SSMZ,soilvariability,MZANDVI,MZsoilproperties}.  
Most of these methods are based on unsupervised learning techniques; specifically, clustering methods such as $k$-means~\cite{MZkmeans} and fuzzy $c$-means~\cite{SSMZ,soilvariability,MZANDVI,MZA}, and principal component analysis (PCA)~\cite{SSMZ,soilvariability,MZsoilproperties}. 
In addition, MZ delineation methods based on supervised learning such as random forests (RFs) and support vector machines (SVMs)~\cite{MZmultilayer,MZsoilproperties} have emerged recently.  

All previous works produce management zones using factors that are directly or indirectly related to crop productivity.
Nevertheless, to the best of our knowledge, no published work has explicitly considered fertilizer responsivity as the main driver for defining MZs.
One of the key objectives of using MZs is to equip farmers with the necessary tools to make informed decisions about crop management, such as determining the appropriate amount of fertilizer required in each zone.
As a consequence, our efforts should be directed toward establishing management zones where all included sites display comparable responsivity to varying fertilizer rates~\cite{yieldpatterns}.
%"If response curves for various inputs such as N are unique for the different yield clusters, then varying input rates may be a cost-effective strategy for corn production on this field. "

Fertilizer responsivity can be characterized using nitrogen fertilizer-yield response (N-response) curves.
These curves exhibit the estimated crop yield values corresponding to a specific field site in response to all admissible fertilizer rates, typically ranging between $0$ and $150$ pounds per acre (lbs/acre)~
% [R1, R2]\footnote{These references are hidden for double-blind review purposes and will be provided after acceptance.}.
\cite{nresponse,rcurves2023}.
The shape of the N-response curve indicates the site's responsiveness to fertilizer, with a flat curve suggesting low responsivity and a steep curve suggesting high responsivity. 
Thus, in this work, we propose an MZ clustering method that accounts for within-field variability of fertilizer responsivity based on approximated N-response curves.

To do this, we derive non-parametric response curves from observed data.
Most response curves are generated based on parametric assumptions using methods such as linear regression or quadratic plateau regression; however, our experience suggests that N-response is much more complex than these models can capture.
Our approach is based on a convolutional neural network (CNN) acting as a regression model to map the covariate factors to crop yield, as suggested in 
% [R3]\footnotemark[1].
\cite{Morales_2023}.
Then, the CNN is used to generate approximated N-response curves across a range of admissible fertilizer rates  
% [R2]\footnotemark[1].
\cite{rcurves2023}.
The distinction in shape between two N-response curves is quantified by measuring the distance between the corresponding transformed curves in a reduced space, calculated using functional principal component analysis (fPCA). 
Thus, determining MZs relies on leveraging the shape dissimilarity of N-response curves as the key distance metric in cluster analysis.

It is worth pointing out that none of the existing methods for determining MZs attempt to provide explanations regarding the behavior of their results. 
This presents a significant limitation, particularly within the context of the growing area of explainable artificial intelligence (XAI). 
Therefore, an approach to determining MZs that is inherently explainable should enable farmers to discern cause-and-effect relationships between their inputs and outputs, enabling a more transparent decision-making process. 
Therefore, we use a \textit{post-hoc} explainability method to facilitate understanding the impact of various covariates on determining the MZ assignment for a given site. 
Specifically, we employ a counterfactual explanation (CFE) method, adapted from a previous work 
% [R2]\footnotemark[1].
\cite{rcurves2023}.

For a specific site within the field, the aim of CFE is to identify the minimal set of covariate factors that, when modified, leads to the response curve of the generated counterfactual sample being assigned to a cluster different from the original one. 
The problem of generating such CFEs is tackled as a multi-objective optimization problem (MOO) and is solved using an approach based on the Non-dominated Sorting Genetic Algorithm II (NSGA-II)~\cite{nsga2}.   
Applying this process across all sites associated with a specific management zone enables the computation of global relevance scores. 
These scores facilitate identifying covariates with the most significant impact on cluster membership assignment. 

Our specific contributions are summarized as follows. 
First, we present the first MZ generation approach that accounts for within-field variability of fertilizer responsivity based on neural network-generated N-response curves derived from observed data.
Second, we show how to apply a \textit{post-hoc} explainability method that generates CFEs to reveal the influence of covariate factors on MZ assignments.

\section{Related Work} \label{sec:related}

Several studies in PA have tackled the challenge of determining MZs, employing a spectrum of techniques ranging from traditional statistical methods to machine learning algorithms.
For instance, Georgi \textit{et al.}~\cite{automaticMZ} developed a cost-effective segmentation algorithm using multi-spectral satellite data.
They generate normalized difference vegetation index (NDVI) maps and apply morphological operations to identify homogeneous regions.
The NDVI values found within the processed maps are then classified into four quantiles, which are used to delineate four MZs.
Two assumptions are made: a direct correlation exists between NDVI and crop yield, and MZs can be determined based on expected productivity.  

Gallardo-Romero \textit{et al.}~\cite{MZmultilayer} presented work based on similar assumptions, using vegetation indices extracted from multispectral images, along with available yield maps from previous growing seasons.
The study categorizes yield values into three classes, assuming that each category corresponds to a distinct MZ. 
To predict productivity levels (i.e., MZs), the authors employ an ensemble of four machine learning algorithms: classification and regression trees (CART), RF, gradient boosting trees (GBT), and SVM. 

\begin{figure}[!t]
    \centering
    \includegraphics[width=3.5cm]{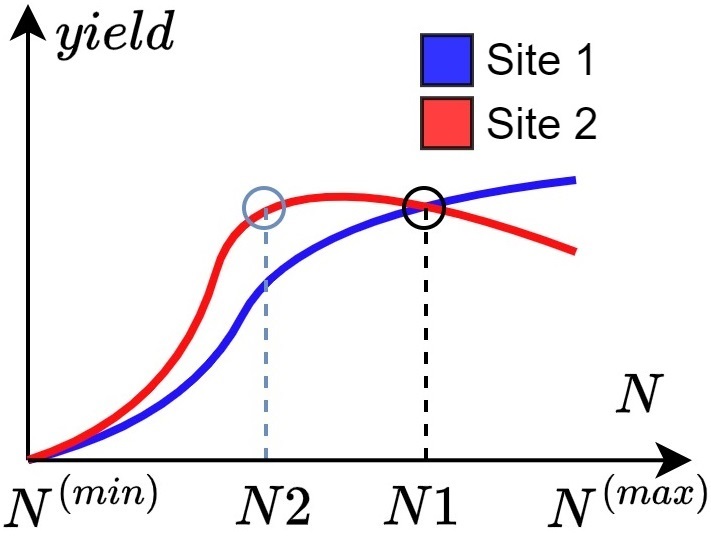}
    % \vspace{-1ex}
    \caption{N-response curve comparison for two sites with the same predicted yield value given a treatment of $N1$ lbs/ac.}
    % \vspace{-1ex}
    \label{fig:curves_comp}
\end{figure}

To motivate our method, we argue that the similarity in estimated yield values for different sites under specific conditions does not imply that these sites would exhibit similar responses under all admissible conditions.
% For instance, consider the case that, in a previous growing season, two sites produced comparable yield values and, as such, were classified within the same management zone.
For instance, take two sites whose predicted yield values are the same when using the same treatment and, as such, are classified within the same management zone.
Assume we know their N-response curves, as depicted in Fig.~\ref{fig:curves_comp}.
Despite both zones producing the same yield when using $N1$ lbs/ac, notice that they react to differing amounts of nitrogen differently. 
In fact, the second zone can achieve the same yield with less nitrogen ($N2$)!
%what is more, the second zone achieves the same yield value using a lower N rate.  

Instead of focusing on predicting productivity levels, other methods have presented techniques to cluster homogeneous regions based on the characteristics of the field.
Fridgen \textit{et al.}~\cite{MZA} introduced the Management Zone Analyst (MZA), a tool that utilizes a fuzzy classification procedure. 
MZA also provides performance indices such as the fuzziness performance index (FPI) and normalized classification entropy (NCE) to aid in determining the optimal number of clusters for MZs. 
This tool has been applied with other clustering methods for determining MZs based on soil traits, NDVI data, and multi-year crop yields \cite{MZANDVI}.

Fuzzy c-means clustering, often paired with PCA, has been used widely for determining site-specific MZs~\cite{SSMZ,soilvariability,fuzzy3}.
These methods take into account the multi-dimensional nature of soil variability, allowing for more meaningful zone delineation.
Other approaches involve integrating fuzzy c-means and PCA with other machine learning algorithms, such as RF.
For example, Maleki \textit{et al.}~\cite{MZsoilproperties} trained an RF model to predict various soil properties (e.g., pH, sand content, and electroconductivity) given environmental covariates (e.g., vegetation indices, geology maps, and water quality parameters).
The RF model was used for feature selection to remove unimportant environmental covariates.
PCA was applied to the reduced features, and the MZs were identified using fuzzy c-means.

Other previous methods overlooked the importance of the explainability aspect concerning the resulting MZs.
Addressing this gap, our work tackles the issue by incorporating counterfactual explanation analysis, aiming to uncover causal dependencies between inputs and outputs.
Nevertheless, there has been limited focus on CFE methods in the context of regression tasks.
Classification-based CFE methods aim to produce counterfactual samples with predicted class labels that differ from the original ones~\cite{CFEclass}. %~\cite{CFEclass,CFEMOO}
In this context, identifying features of higher relevance involves recognizing those undergoing more frequent changes during CFE generation~\cite{CFErelev}.

Based on this concept, we proposed a CFE method 
% [R2]\footnotemark[1] 
\cite{rcurves2023} 
that, given a response curve generated for a selected input feature (known as the ``active feature"), allows for the identification of the remaining system variables (known as the ``passive features") with the highest relevance on the curve's shape.
This method diverges from classic sensitivity analysis or saliency analysis, typically employed to explore how various values of a set of independent variables influence the response variable. 
In contrast, our approach delves into understanding how different combinations of the passive features influence the response variable across the entire range of admissible values of the active feature, without assuming independence among the features.
In this work, we adapt this method to understand the impact of passive features on the MZ membership of a field site. 
Thus, a new objective function will be presented.

\section{Materials and Methods}

\subsection{Datasets}
\label{sec:data}

We utilized two datasets for early-yield prediction acquired from two winter wheat dryland fields called ``Field A" and ``Field B". 
These datasets were compiled and discussed in a previous work
% [R3]\footnotemark[1].
\cite{Morales_2023}.
The problem of predicting crop yield is formulated as a regression task, with a set of eight covariates collected in March serving as the explanatory variables:
\begin{enumerate}
   \item $N$: Nitrogen rate (lb/ac).
   \item $S$: Topographic slope (degrees).
   \item $E$: Topographic elevation (\SI{}{m}).
   \item $TPI$: Topographic position index.
   \item $A$: Topographic aspect (radians).
   \item $P$: Precipitation from the prior year (\SI{}{mm}).
   \item $VV$ and $VH$: Backscattering coefficients derived from synthetic aperture radar (SAR) images from Sentinel-I.
\end{enumerate}
The dependent variable is the harvested yield in bushels per acre (bu/ac), recorded post-harvest in August. 
Therefore, the March-acquired data serve as predictors for the yield outcomes obtained in August.
The acquired feature and yield maps can be seen as image rasters, where each pixel or cell represents a region of $10 \times 10\,$\SI{}{\meter}.
Data were collected across three growing seasons for each field (2016, 2018, and 2020).

\subsection{Convolutional Neural Network} \label{sec:CNN}

In
% [R3]\footnotemark[1],
\cite{Morales_2023},
we tackled the yield prediction problem using a two-dimensional (2D) regression model; that is, a model with 2D inputs and 2D outputs.
We designed a 3D--2D CNN architecture called 
% [Hidden name]\footnotemark[1],
Hyper3DNetReg,
which is trained to predict the yield values of all cells within a small spatial neighborhood of a field simultaneously.
Specifically, our model takes as input an image data cube of $5 \times 5$ pixels with $n=8$ channels, where each channel represents a distinct covariate. 
The model then generates a 2D image output of $5 \times 5$ pixels.

Hence, data from each field were partitioned into several $5 \times 5$--pixel patches to create their corresponding training and validation sets.
The training dataset comprised 90\% of the patches extracted from three available years of data, with the remaining 10\% constituting the validation dataset.
Our trained models are field-specific, signifying that they are trained on data from a particular field and employed to predict future yield maps using data from the same crop and the same field.
% For this study, fields A and B were trained under a ``leave-one-year-out" configuration, employing 2016 and 2018 data for training and validation, and 2020 data for testing.
% The trained models were used to generate predicted yield maps for the test years, producing root mean squared errors (RMSEs) of 14.87 and 12.31 for fields A and B, respectively. 

\subsection{N-response Curve Generation}

After the network has been trained and, assuming it captures the underlying causal structure of the problem adequately, it can be employed to produce approximate response curves.
In previous work 
% [R2]\footnotemark[1], 
\cite{rcurves2023},  
we defined a response curve as a tool that allows for the responsivity analysis of a sensitive system to a particular ``active feature". 
In addition, other stimuli that may influence the relationship between the response variable and the active feature were referred to as ``passive features".
In the context of the generation of N-response curves, the active feature corresponds to $N$ (nitrogen rate), while the seven remaining variables represent the set of passive features. 

\begin{figure}[!t]
    \centering
    \includegraphics[width=7.5cm]{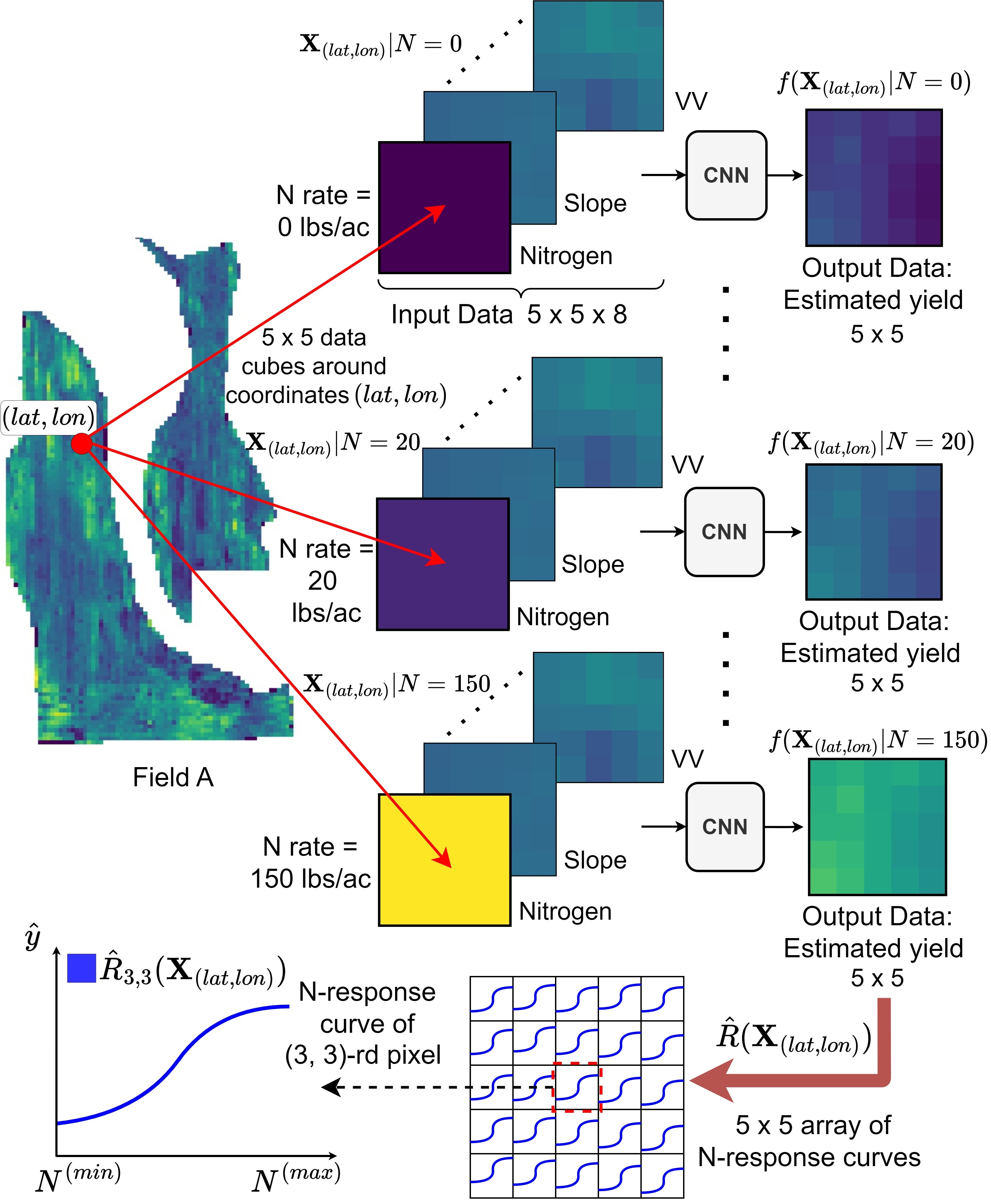}
    % \vspace{-1ex}
    \caption{Generation of a $5 \times 5$ array of N-response curves generated around a field point at coordinates $(lat, lon)$.}
    % \vspace{-1ex}
    \label{fig:rcurves_gen}
\end{figure}

An input data cube is represented as $\mathbf{X} = \{ X_1, \dots, X_n\}$, with $X_1$ corresponding to the $N$ covariate, and the subsequent dimensions aligning with the remaining covariates based on the order outlined in Section~\ref{sec:data}.
Let us denote the trained CNN model as $f(\cdot)$ so that, given an input $\mathbf{X}$, its estimated yield patch is denoted as $\hat{\mathbf{Y}} = f(\mathbf{X})$. 
Consider we produce estimated yield patches for all admissible values of $N$ (bounded by $N^{\min} \le N \le N^{\max}$) and stack them as a data cube $\hat{R}(\textbf{X})$: 
\begin{equation}
    \hat{R}(\textbf{X}) = \{ f(\textbf{X} | N = N^{\min}), \dots , f(\textbf{X} | N = N^{\max})\},
    \label{eq:curvegen}
\end{equation}
as shown in Fig.~\ref{fig:rcurves_gen}. 
As such, the $(i,j)$-th cell of $\hat{R}(\textbf{X})$ (where $1 \leq i, j \leq 5$), denoted as $\hat{R}_{i,j}(\textbf{X})$, represents the approximated response curve corresponding to the $(i,j)$-th cell of input $\mathbf{X}$.

\begin{figure*}[!t]
    \centering
    \includegraphics[width=17.7cm]{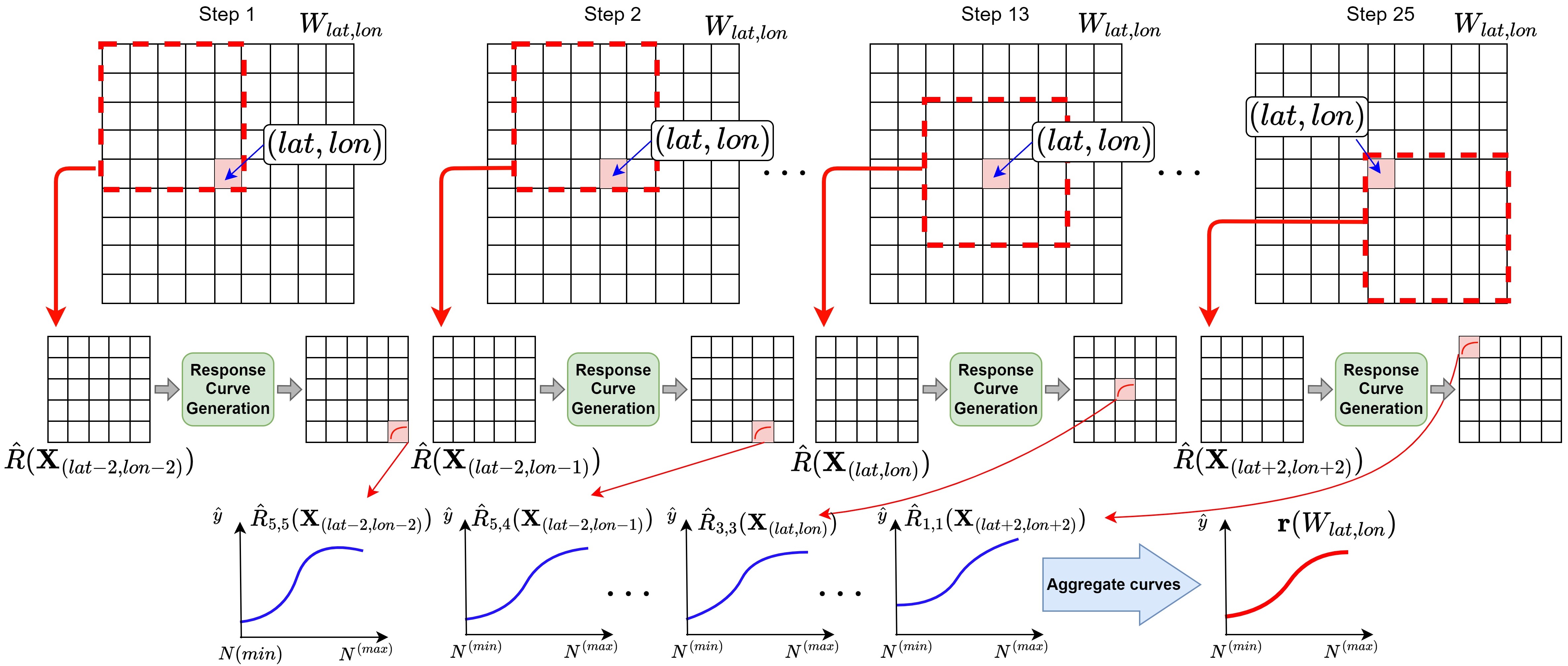}
    % \vspace{-1ex}
    \caption{N-response curves aggregation for a field point at coordinates $(lat, lon)$.}
    % \vspace{-1ex}
    \label{fig:rcurves_agg}
    \vspace{-1ex}
\end{figure*}

The example in Fig.~\ref{fig:rcurves_gen} shows the response curve generation process of all pixels within input $\mathbf{X}_{(lat, lon)}$, which represents the $5 \times 5$--pixel patch generated around coordinates $(lat, lon)$ of the field.   
However, our goal is to generate N-response curves for all sites within the field.
To do so, we move a $5 \times 5$--pixel sliding window throughout the entire field.
Note that this approach results in overlapping predicted yield patches for consecutive points.
Therefore, results obtained from neighboring points of the field must be aggregated. 

Fig.~\ref{fig:rcurves_agg} depicts the aggregation process of N-response curves obtained for a field point at coordinates $(lat, lon)$.
This process involves considering all valid neighboring $5 \times 5$--pixel patches; i.e., patches whose centers are located within the field and that contain the point at $(lat, lon)$, highlighted in red.
The $9 \times 9$ window generated around the field point at $(lat, lon)$ is denoted as $W_{lat, lon}$.
For each of the valid patches in $W_{lat, lon}$, a $5 \times 5$ array of N-response curves is generated using Eq.~\ref{eq:curvegen}.
Then, the N-response curves corresponding to the field point at $(lat, lon)$ are averaged, yielding a singular approximated N-response $\mathbf{r}(W_{lat, lon})$. 
This averaging process alleviates noisy outcomes and produces smoothed curves.

In Sec.~\ref{sec:intro}, we stated that the fertilizer responsivity of a given site is characterized by the shape of its N-response curve.
As such, when comparing the shape of two or more N-response curves, the focus is not on their absolute estimated yield values.
Hence, any vertical shifts are eliminated to obtain the aligned approximate N-response curve $\tilde{\mathbf{r}}(W_{lat, lon})$ as follows:
\[
\tilde{\mathbf{r}}(W_{lat, lon}) = \mathbf{r}(W_{lat, lon}) - \min({\mathbf{r}}(W_{lat, lon})).
\]

\subsection{Functional Principal Component Analysis} \label{sec:fPCA}

We compute the set $\mathbf{R}$ comprising the aligned approximate N-response curves generated for all sites within the field.
$\mathbf{R}$ constitutes a set of functional data whose samples are approximated N-response curves.
Thus, fPCA can be applied to $\mathbf{R}$ to establish a distance metric conveying the difference in shape between N-response curves, as suggested in  
% [R2]\footnotemark[1].
\cite{rcurves2023}. 

Functional Principal Component Analysis extends traditional PCA to analyze and represent variability in functional data~\cite{FDA}.
As such, an N-response curve can be expressed as a linear combination of \textit{functional} principal component (fPCs). 
Each fPC encapsulates a unique curve pattern, implying that curves with distinct shapes will be encoded using different fPC values.
In this work, we suggest approximating an N-response curve using $K=3$ fPCs, a choice justified by their ability to explain at least 99.5\% of the variance of fields A and B.  
Thus, the proposed distance metric between curves $\mathbf{r}_1$ and $\mathbf{r}_2$ is:
\begin{equation}
    d(\mathbf{r}_1, \mathbf{r}_2) = \sqrt{ \sum_{k=1}^{K} \left( v_k(\mathbf{r}_1) - v_k(\mathbf{r}_2) \right) ^2},
    \label{eq:dist}
\end{equation}
where $v_k(\mathbf{r}_j)$ is the value of the $k$-th principal component obtained after transforming the curve $\mathbf{r}_j$.

\subsection{Management Zone Clustering} \label{sec:clustering}

Using fuzzy c-means has become a prevalent approach in management zone delineation methods~\cite{SSMZ,soilvariability,fuzzy3}.
In fuzzy c-means, each data point is assigned a membership score indicating the extent to which it belongs to a specific cluster. 
A cluster centroid is computed as the mean of all data points, weighted by their respective cluster membership values.

We propose to cluster all field points based on their fertilizer responsivity so that each cluster corresponds to a distinct MZ.
The process involves generating aligned approximate N-response curves for all field sites, followed by their transformation into a reduced three-dimensional (3D) space through fPCA.
Hence, the difference in fertilizer responsivity between curves (i.e., the difference in curve shape) is conveyed by their Euclidean distance in the transformed space.
Therefore, the fertilizer responsivity distance (Eq.~\ref{eq:dist}) serves as the distance metric for the fuzzy c-means algorithm.

Some approaches utilize indices such as the silhouette score, fuzziness performance index, and normalized classification entropy to determine the optimal number of clusters~\cite{MZANDVI,MZA}. 
However, these indices might face challenges in situations where clusters lack clear separation, as observed in the present context.
Recall that all data points for clustering belong to the same field, leading to gradual changes in soil variability and, as a consequence, gradual changes in fertilizer responsivity.

In addition, in PA, it is a common practice to specify between three and five MZs~\cite{MZmultilayer}.
The decision to use up to five MZs is often influenced by practical considerations, such as the limitations of variable rate application machinery and the complexity of the field.
For instance, using more than five zones may entail intricate zone boundaries, posing challenges for certain variable rate technologies to distinguish between closely situated zones.
Following this convention, we chose through visual inspection a cluster count that minimizes the creation of redundant or highly variable MZs.
%This inspection is carried on by visual analysis of the clustered curves.
In future work, we will design statistical tests that will determine if the responsivity of multiple clusters of functional data is significantly different.

\subsection{Counterfactual Explanations for Management Zones} \label{sec:CFE}

% Unlike previous approaches, we seek to provide explanations about the resulting management zones.
Consider a field point at coordinates $(lat, lon)$; the $9 \times 9$--pixel window generated around it, $W_{lat, lon}$; and the aligned N-response curve generated for the center of this window, $\tilde{\mathbf{r}}(W_{lat, lon})$.
Let $W'_{lat, lon}$ represents a counterfactual explanation of $W_{lat, lon}$ and let $\tilde{\mathbf{r}}(W'_{lat, lon})$ be its corresponding aligned N-response curve.
The CFE $W'_{lat, lon}$ is generated by introducing perturbations to the original set of passive features in $W_{lat, lon}$, thereby altering the cluster membership of the resulting N-response curve $\tilde{\mathbf{r}}(W'_{lat, lon})$.
Furthermore, we seek to identify the minimal set of passive features requiring perturbation for a shift in cluster membership, which focuses on identifying features with the highest impact on this alteration.

This problem is tackled as an MOO problem.
To solve it, we adapt the method presented in
% [R2]\footnotemark[1].
\cite{rcurves2023}. 
Originally, this method was proposed to minimize three competing objectives using NSGA-II~\cite{nsga2}. 
Its primary use is to identify the minimal set of passive features requiring modification, ensuring that the distance between the responsivity of the counterfactual explanation (CFE) and that of the original samples exceeds a predefined hyperparameter threshold.
Note that this approach does not account for management zone membership and is used to analyze the overall field behavior.

Specifically, our MOO problem is defined as: %the minimization of the following objectives:
\begin{align*}
\min_{W'_{lat, lon}} ( g_1(W_{lat, lon}, W'_{lat, lon}), &g_2(W_{lat, lon}, W'_{lat, lon}), \\
& g_3(W_{lat, lon}, W'_{lat, lon}) ).
\end{align*}
The first objective differs from the first objective presented in 
% [R2]\footnotemark[1],
\cite{rcurves2023}, 
and aims for a shift in cluster membership, such that:
\begin{equation*}
    g_1(W, W') = 
    \begin{cases}
      - 1, & \begin{aligned}[t]
        &\text{if $\texttt{cl}(\tilde{\mathbf{r}}(W')) \neq \texttt{cl}(\tilde{\mathbf{r}}(W)) \wedge$} \\
        &\text{$\;\;\;\; \texttt{m}(\tilde{\mathbf{r}}(W')) > \epsilon$}
      \end{aligned}\\
      0, & \text{otherwise},
    \end{cases} 
    \label{eq:g1}
\end{equation*}
where $\texttt{cl}(\tilde{\mathbf{r}}(W'))$ is the cluster assignment for a curve $\tilde{\mathbf{r}}(W')$ after undergoing transformation via fPCA, while $\texttt{m}(\tilde{\mathbf{r}}(W')) > \epsilon$ denotes the membership score that $\tilde{\mathbf{r}}(W')$ belongs to the assigned cluster.
Recall that a field point is assigned a membership score for each possible cluster by fuzzy c-means.

Notice that the objective is to alter the cluster membership so that the membership score assigned to the new cluster is greater than $\epsilon$, which is a tunable hyperparameter.
The reasoning behind the decision is as follows.
Suppose a curve $\tilde{\mathbf{r}}(W)$ is assigned to a certain cluster with low membership; then, this curve would be susceptible to changes in cluster membership when subjected to minor random perturbations.
As a result, such changes may not be informative or representative of the behavior of the remaining curves within the same cluster.
To avoid this situation, we enforce that $\texttt{m}(\tilde{\mathbf{r}}(W')) > \epsilon$ (e.g. $\epsilon >0.8$) during CFE generation. 
This ensures that the CFE is assigned to a different cluster with high confidence, making the necessary changes more meaningful.

The second objective is related to minimizing the number of modified features:
\[
    g_2(W, W') = || W - W' ||_0,  
    \label{eq:g2}
\]
where $|| \cdot ||_0$ represents the $L_0$ norm.
Finally, the third objective accounts minimizing the distance between the original sample and its CFE.
% This is achieved through the utilization of the Gower distance, chosen due to the real-valued nature of the covariates in this study, each exhibiting distinct ranges of values:
Considering the real-valued nature of the covariates in this study, each exhibiting distinct ranges of values, the distance is calculated as:
\[
    g_3(W, W') = \frac{1}{n} \sum_{s=1}^n \frac{1}{r_s} | W^{(s)} - W'^{(s)}|, 
    \label{eq:g3}
\]
where $W^{(s)}$ is the $s$-th covariate, and $r_s$ its range of values.

This MOO problem is solved using NSGA-II with a population size of $T_0$ CFE candidate solutions. 
NSGA-II outputs a set of $T$ Pareto-optimal solutions ($T \leq T_0$).
These solutions are non-dominated, meaning that no other solution in the set is superior in all objectives simultaneously.
Hence, our selection criteria for a solution from the Pareto set involve choosing solutions with the minimum $g_1$ value, followed by those with the lowest $g_2$ value, and finally selecting solutions with the minimum $g_3$ value.

Furthermore, we calculate local and global explanations for the generated results.
Given a field point, a local explanation identifies the passive features with the greatest impact on its management zone membership.
As such, a local explanation for a field point at coordinates $(lat, lon)$, denoted as $\alpha_{lat, lon}$ is given by the set of features that were altered during the generation of the counterfactual sample:
\[
\alpha_{lat, lon} = \{ s \; | \; W_{lat, lon}^{(s)} \neq W'^{(s)}_{lat, lon} \}.
\]

\begin{figure}[!t]
    \centering
    \begin{subfigure}{0.49\columnwidth}
        \centering
        \includegraphics[width=\linewidth]{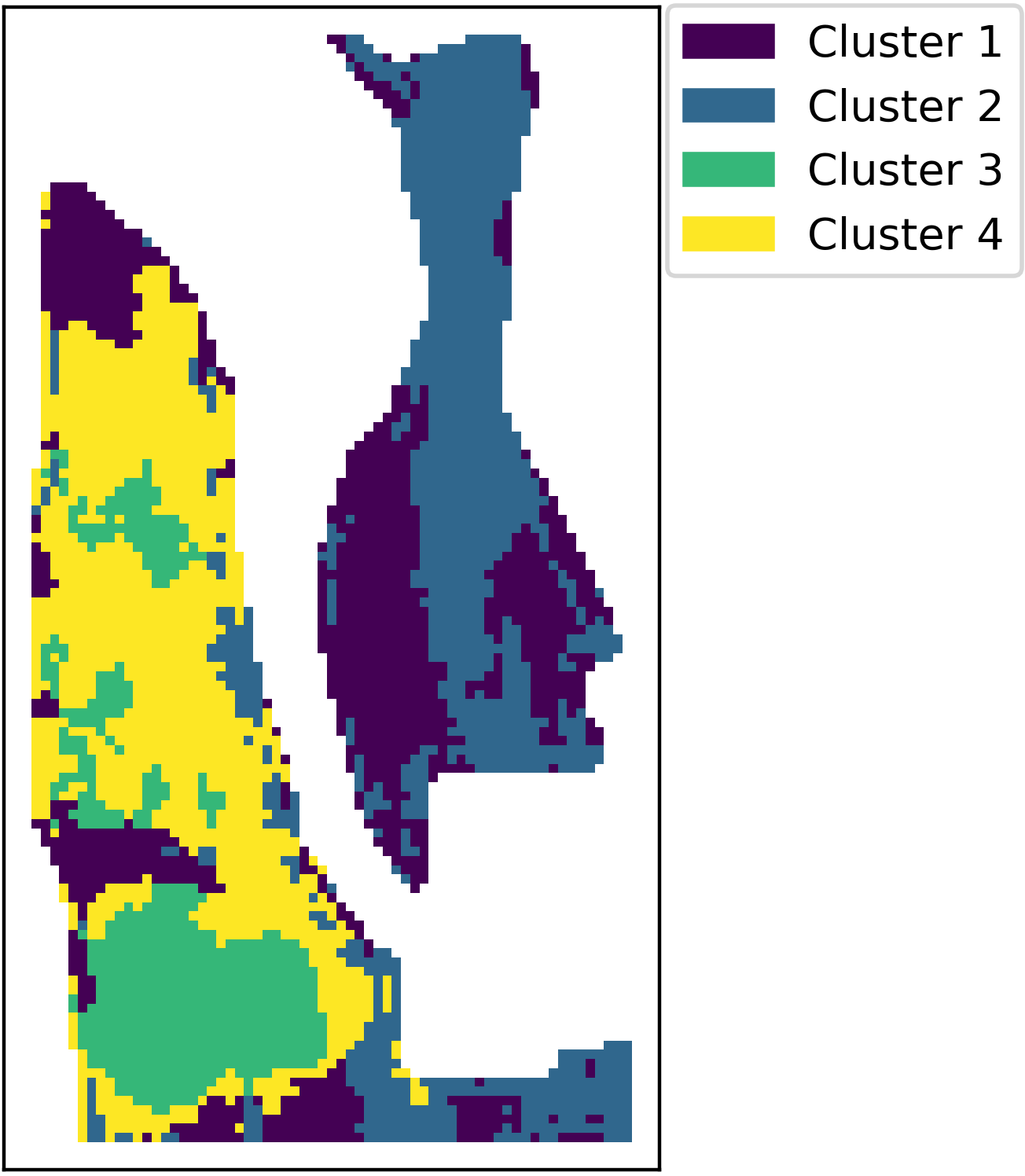} 
    \vspace{-3ex}
        \caption{}
        \label{fig:image_a}
    \end{subfigure}
    \begin{subfigure}{0.4\columnwidth}
        \centering
        \includegraphics[width=\columnwidth]{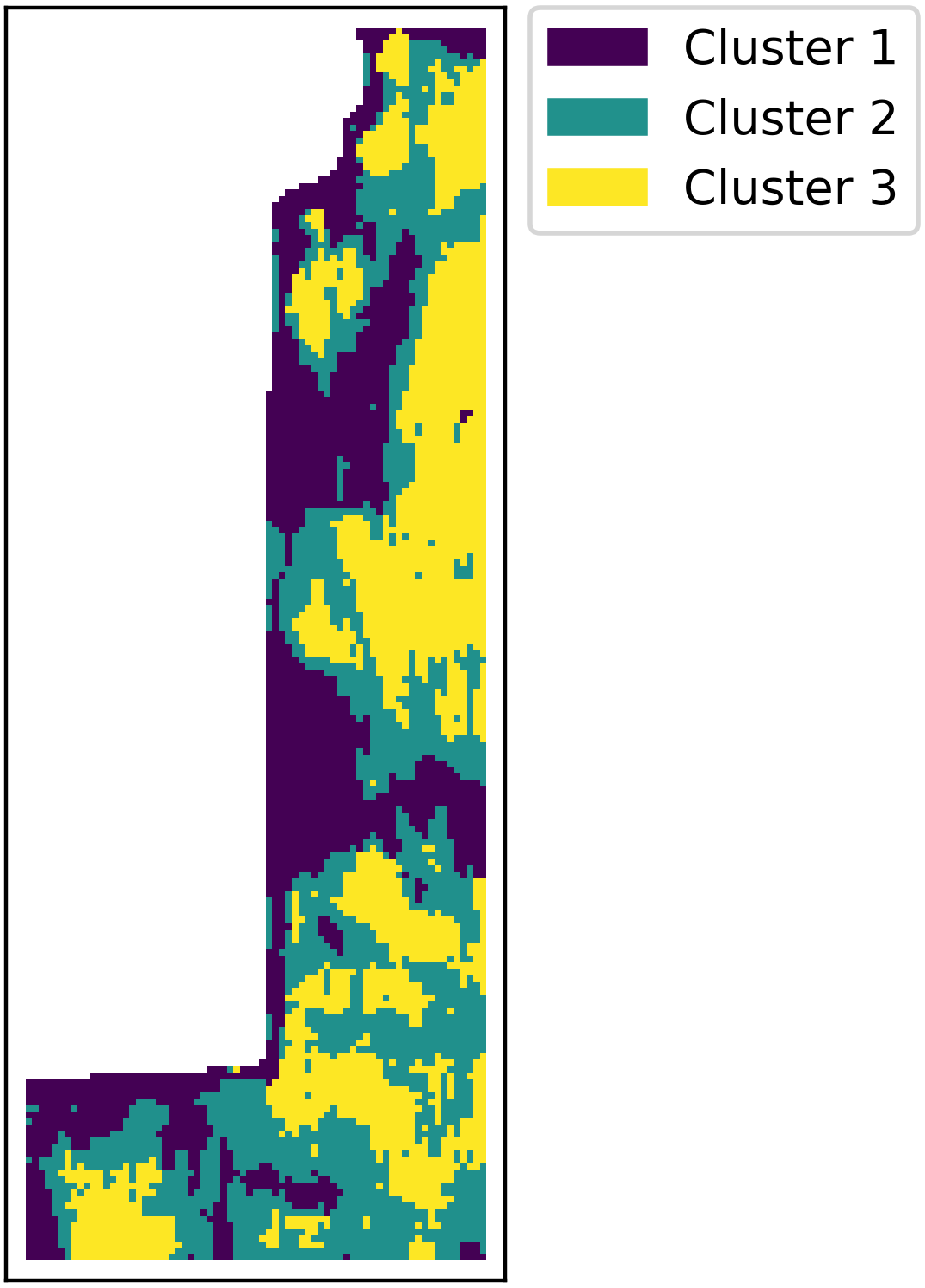} 
    \vspace{-3ex}
        \caption{}
        \label{fig:image_b}
    \end{subfigure}
    \vspace{-1ex}
    \caption{Management zones for (a) Field A and (b) Field B.}
    \vspace{-1ex}
    \label{fig:clustered}
\end{figure}

In contrast, global explanations are generated for \textit{each MZ} and convey the relevance of passive features at the MZ level.  
When analyzing the $z$-th MZ, the individual feature relevance of the $s$-th passive feature is calculated as follows:
\[
r_z^{(s)} = \frac{1}{|\mathbf{C}_z|} \sum_{(lat, long)\, \in \, \mathbf{C}_z}^{|\mathbf{C}_z|} \mathbb{I}_{s \, \in \, \alpha_{lat, lon}},
\]
where $\mathbf{C}_z$ represents the set of field points that have been clustered into the $z$-th MZ.
Thus, $r_z^{(s)}$ is the ratio of times that the $s$-th passive feature was modified when generating CFEs for field points in $\mathbf{C}_z$.

Finally, we emphasize that, in this work, we did not assume mutual independence among the covariate variables.
During the generation of CFEs, more than one passive feature may be modified simultaneously. 
This suggests that certain combinations of features are more effective than modifying individual features in isolation.
Therefore, to analyze which features react together and identify the most effective combinations, we report the five most frequently occurring feature combinations.

\section{Experimental Results} \label{sec:results}

\begin{figure}[t]
    \centering
    \includegraphics[width=\columnwidth]{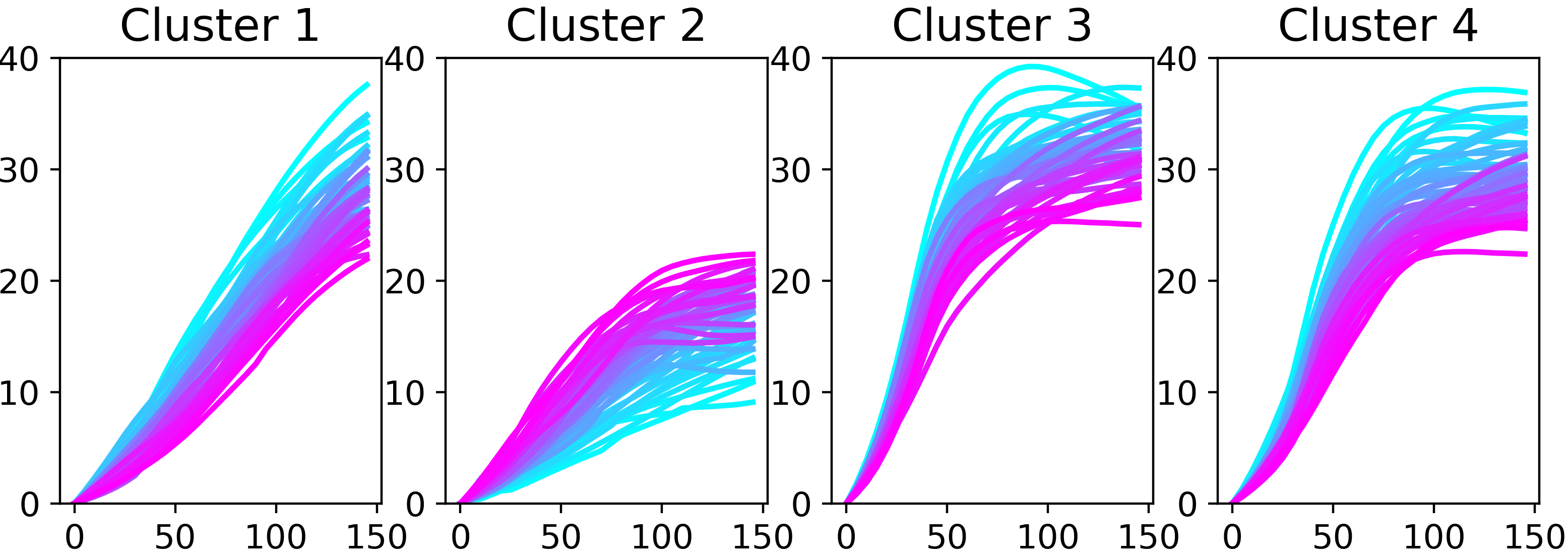}
    % \vspace{-1ex}
    \caption{Aligned approximated N-response curves corresponding to each management zone generated for Field A.}
    % \vspace{-1ex}
    \label{fig:rcurves_fieldA}
    % \vspace{-1ex}
\end{figure}

\begin{figure}[t]
    \centering
    \includegraphics[width=8cm]{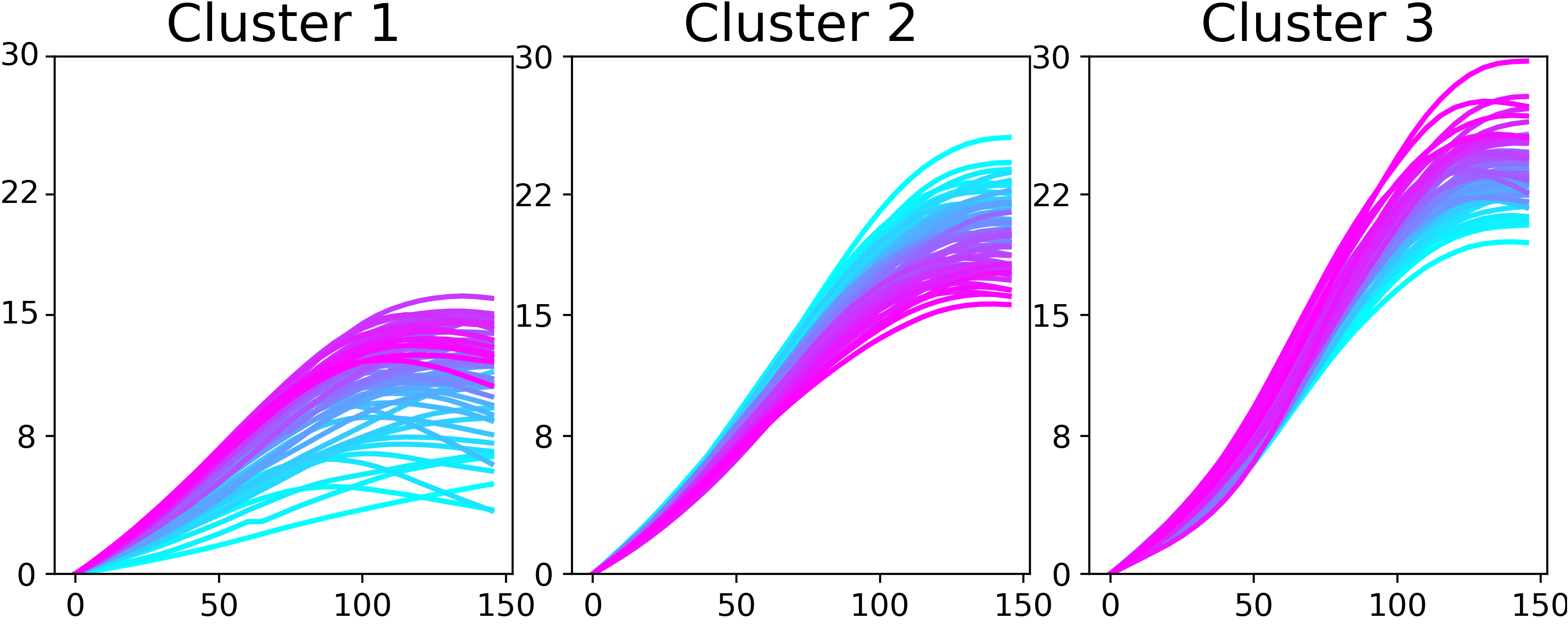}
    % \vspace{-1ex}
    \caption{Aligned approximated N-response curves corresponding to each management zone generated for Field B.}
    % \vspace{-1ex}
    \label{fig:rcurves_fieldB}
    % \vspace{-1ex}
\end{figure}

We evaluated our MZ clustering method outlined in Sec~\ref{sec:clustering} on Fields A and B. 
For Field A, we decided to generate four MZs (i.e., four clusters).
Conversely, for Field B, we chose to generate three MZs.
This decision is justified by the fact that Field B is more homogeneous than Field A, thus their corresponding N-response curves show less variability. %, as we will discuss in Sec.~\ref{discussion}.
The MZs obtained for Fields A and B are shown in Figs.\ref{fig:image_a} and~\ref{fig:image_b}, respectively.
In addition, Fig.\ref{fig:rcurves_fieldA} and Fig.\ref{fig:rcurves_fieldB} show fifty aligned approximate response curves selected randomly from each MZ obtained from each field.
The implementation code is available at \url{https://github.com/NISL-MSU/ManagementZonesCFE}.

\begin{figure}[!t]
    \centering
    \includegraphics[width=8cm]{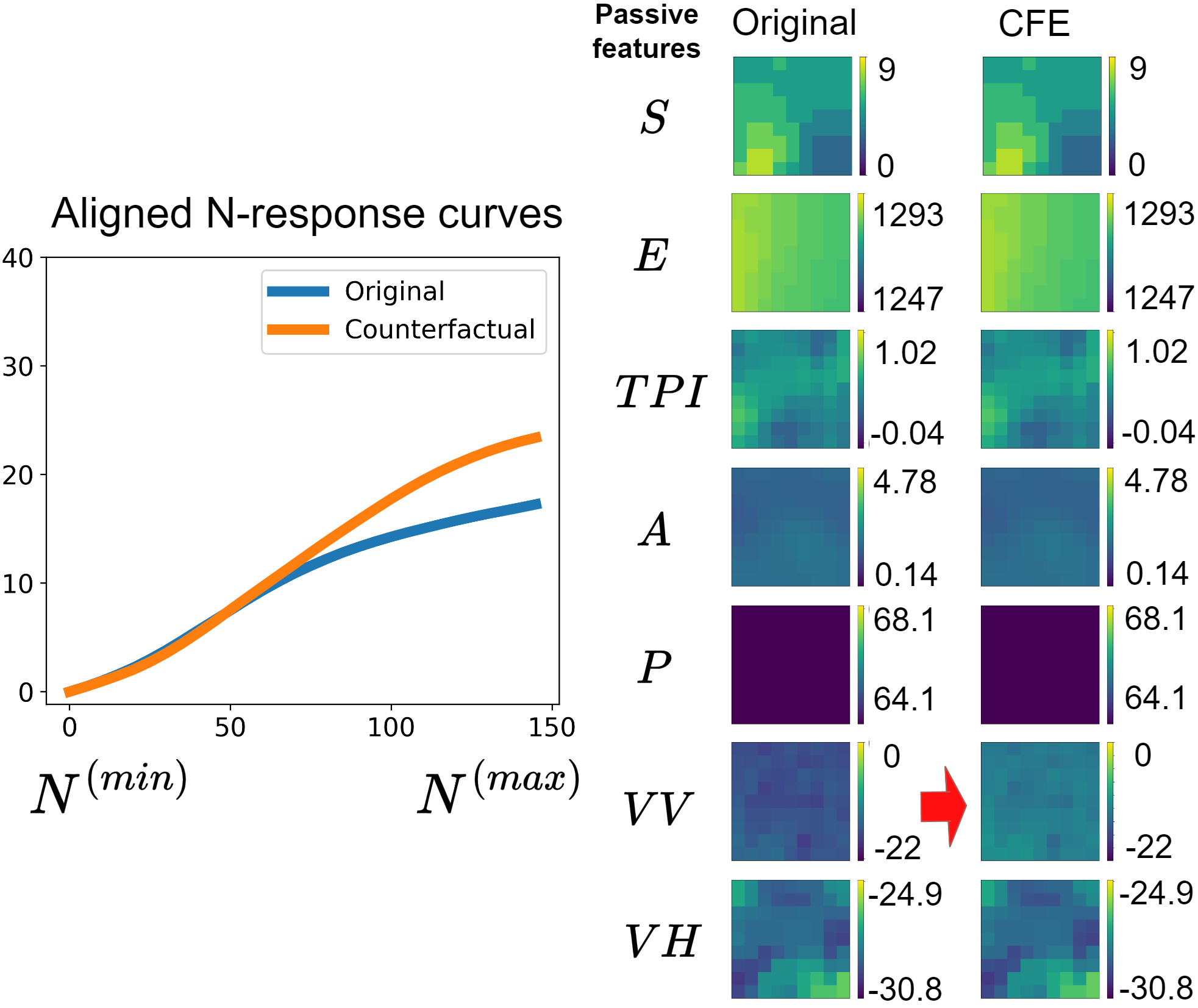}
    % \vspace{-1ex}
    \caption{Example of a counterfactual N-response curve generated for a field point of management zone ``2" of Field A.}
    % \vspace{-1ex}
    \label{fig:CFE_example}
    % \vspace{-1ex}
\end{figure}

Furthermore, we applied our CFE analysis method to all field points of all MZs created for Fields A and B.
For NSGA-II, we used a population size of $T_0=50$ and $100$ iterations.
For objective $g_1(\cdot)$, we selected $\epsilon = 0.8$.
For example, Fig.~\ref{fig:CFE_example} illustrates a counterfactual N-response curve generated for a site of Field A that was initially clustered into MZ ``2."
This CFE exhibits an increase in the $VV$ variable, resulting in an N-response curve with higher responsivity and a shift in cluster membership to "1."
We conducted experiments with alternative $\epsilon$ values, such as $0.85$ and $0.9$, and observed similar results, although they led to slower convergence rates. 
These findings are not included in this paper due to space limitations.

\begin{figure}[t]
    \centering
    \includegraphics[width=6cm]{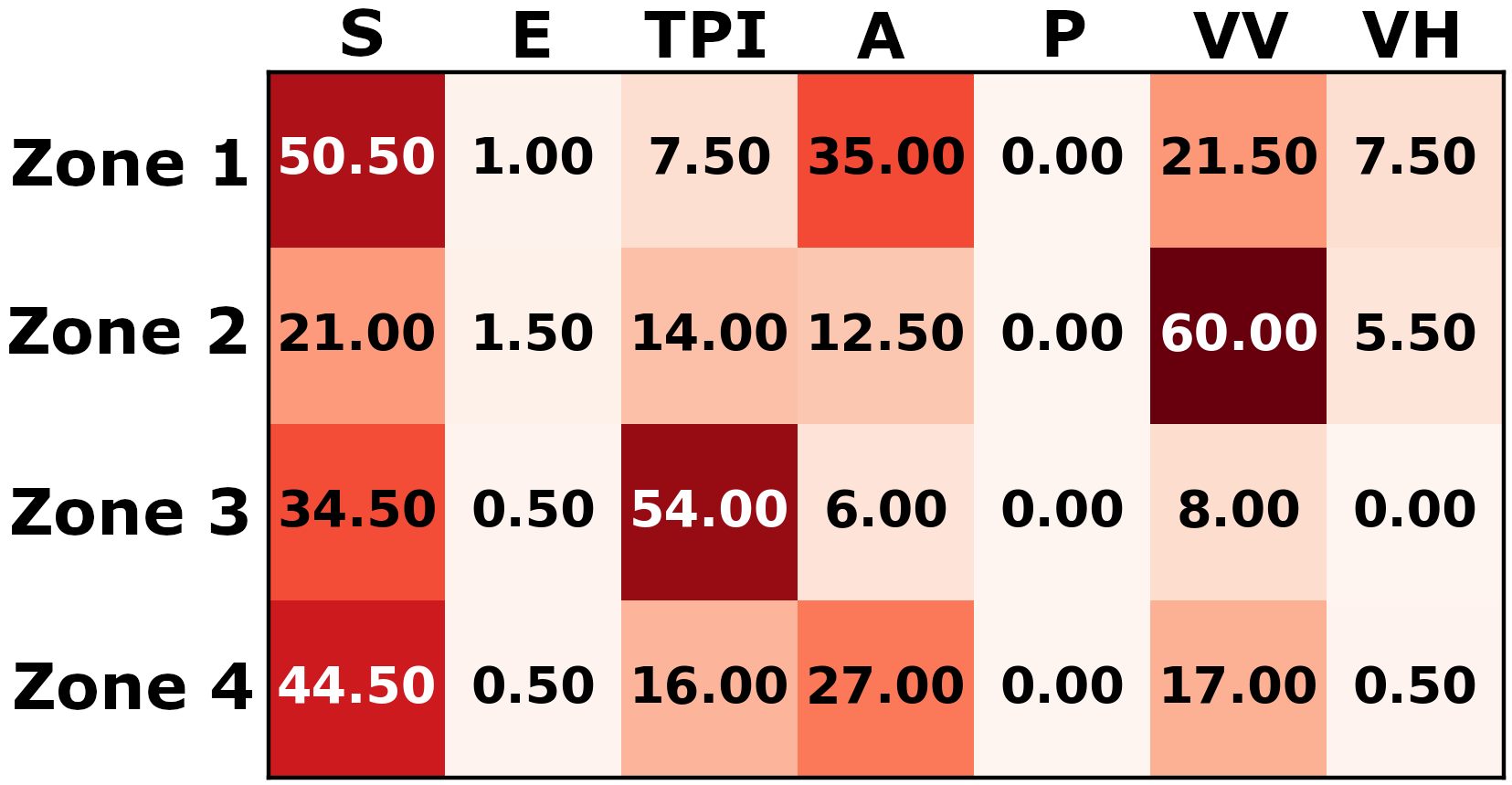}
    % \vspace{-1ex}
    \caption{Individual feature relevance values of Field A.}
    % \vspace{-1ex}
    \label{fig:CFE_fieldA}
\end{figure}

\begin{figure}[t]
    \centering
    \includegraphics[width=6cm]{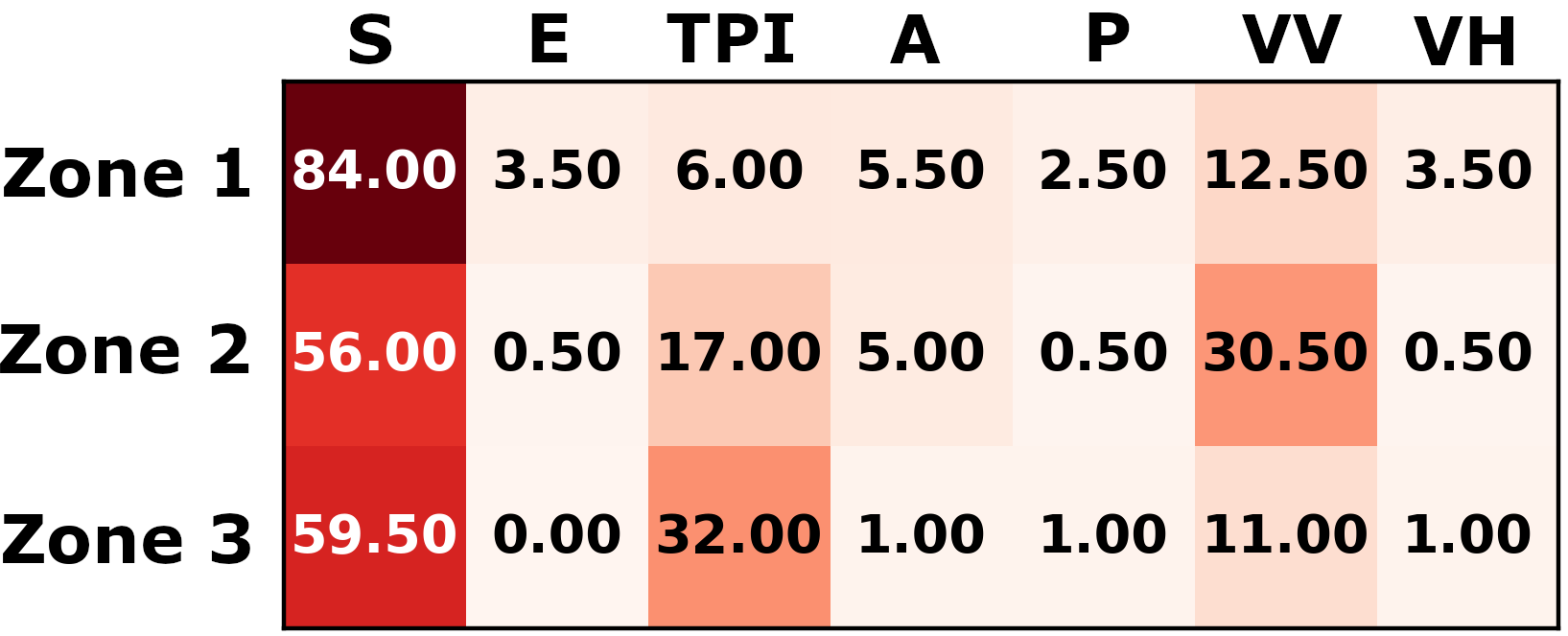}
    % \vspace{-1ex}
    \caption{Individual feature relevance values of Field B.}
    % \vspace{-1ex}
    \label{fig:CFE_fieldB}
\end{figure}

Finally, we show the calculated individual feature relevance values obtained for all passive features of Fields A and B in Fig.~\ref{fig:CFE_fieldA} and Fig.~\ref{fig:CFE_fieldB}, respectively.
In addition, Tables\ref{tab:fieldA_comb} and \ref{tab:fieldB_comb} list the five most repeated feature combinations found during the CFE analysis of each MZ.

\begin{table}[t]
    \centering
    \large
    \caption{Top-five feature combinations -- Field A}
    \label{tab:fieldA_comb}
    \vspace{-1ex}
    \resizebox{\columnwidth}{!}{
    \def\arraystretch{1.2}
    \begin{tabular}{|c|cc|cc|cc|cc|}
    \hline
    Zone & \multicolumn{2}{c|}{1} & \multicolumn{2}{c|}{2} & \multicolumn{2}{c|}{3} & \multicolumn{2}{c|}{4} \\ \hline
    \textbf{\#} & \multicolumn{1}{c|}{Comb.} & \% Rep. & \multicolumn{1}{c|}{Comb.} & \% Rep. & \multicolumn{1}{c|}{Comb.} & \% Rep. & \multicolumn{1}{c|}{Comb.} & \% Rep. \\ \hline
    \textbf{1} & \multicolumn{1}{c|}{{[}S{]}} & 39.5 & \multicolumn{1}{c|}{{[}VV{]}} & 50 & \multicolumn{1}{c|}{{[}TPI{]}} & 53 & \multicolumn{1}{c|}{{[}S{]}} & 41 \\ \hline
    \textbf{2} & \multicolumn{1}{c|}{{[}A{]}} & 22 & \multicolumn{1}{c|}{{[}S{]}} & 16 & \multicolumn{1}{c|}{{[}S{]}} & 33 & \multicolumn{1}{c|}{{[}A{]}} & 23.5 \\ \hline
    \textbf{3} & \multicolumn{1}{c|}{{[}VV{]}} & 10.5 & \multicolumn{1}{c|}{{[}A{]}} & 8 & \multicolumn{1}{c|}{{[}VV{]}} & 5.5 & \multicolumn{1}{c|}{{[}VV{]}} & 16.5 \\ \hline
    \textbf{4} & \multicolumn{1}{c|}{{[}S, A{]}} & 5 & \multicolumn{1}{c|}{{[}TPI{]}} & 7 & \multicolumn{1}{c|}{{[}A{]}} & 5.5 & \multicolumn{1}{c|}{{[}TPI{]}} & 13 \\ \hline
    \textbf{5} & \multicolumn{1}{c|}{{[}A, VV{]}} & 3.5 & \multicolumn{1}{c|}{{[}VH{]}} & 4 & \multicolumn{1}{c|}{{[}S, VV{]}} & 1.5 & \multicolumn{1}{c|}{{[}S, A{]}} & 2 \\ \hline
    \end{tabular}
    }
    % \vspace{-1ex}
\end{table}

\begin{table}[t]
    \centering
    \small %\scriptsize
    \caption{Top-five feature combinations -- Field B}
    \label{tab:fieldB_comb}
    \vspace{-1ex}
    \resizebox{\columnwidth}{!}{
    \def\arraystretch{1.2}
    \begin{tabular}{|c|cc|cc|cc|}
    \hline
    Zone & \multicolumn{2}{c|}{1} & \multicolumn{2}{c|}{2} & \multicolumn{2}{c|}{3} \\ \hline
    \textbf{\#} & \multicolumn{1}{c|}{Comb.} & \% Rep. & \multicolumn{1}{c|}{Comb.} & \% Rep. & \multicolumn{1}{c|}{Comb.} & \% Rep. \\ \hline
    \textbf{1} & \multicolumn{1}{c|}{{[}S{]}} & 66.5 & \multicolumn{1}{c|}{{[}S{]}} & 49 & \multicolumn{1}{c|}{{[}S{]}} & 55.5 \\ \hline
    \textbf{2} & \multicolumn{1}{c|}{{[}S, VV{]}} & 11 & \multicolumn{1}{c|}{{[}VV{]}} & 23 & \multicolumn{1}{c|}{{[}TPI{]}} & 26.5 \\ \hline
    \textbf{3} & \multicolumn{1}{c|}{{[}TPI{]}} & 6 & \multicolumn{1}{c|}{{[}TPI{]}} & 13.5 & \multicolumn{1}{c|}{{[}VV{]}} & 9 \\ \hline
    \textbf{4} & \multicolumn{1}{c|}{{[}S, A{]}} & 3 & \multicolumn{1}{c|}{{[}S, VV{]}} & 4.5 & \multicolumn{1}{c|}{{[}S, TPI{]}} & 3 \\ \hline
    \textbf{5} & \multicolumn{1}{c|}{{[}S, TPI{]}} & 1.5 & \multicolumn{1}{c|}{{[}A{]}} & 3 & \multicolumn{1}{c|}{{[}VV, TPI{]}} & 1.5 \\ \hline
    \end{tabular}%
    }
    % \vspace{-1ex}
\end{table}

\section{Discussion} \label{discussion}

 Our method involves using CNNs for the automated generation of approximated N-response curves, which are employed to describe the fertilizer responsivity at different locations in the field.
 Fig.~\ref{fig:rcurves_fieldA} and Fig.~\ref{fig:rcurves_fieldB} show a subset of these non-parametric curves, demonstrating a behavior aligned with agronomic expectations; i.e., sigmoid-\textit{like} curves that capture an apparent yield loss after reaching a saturation point. %~\cite{watkins}. 

 These figures confirm that our MZ determination method organized the generated N-response curves effectively into clusters with consistent curve shapes within each cluster. 
 The observed variation in curve shapes across different clusters highlights different patterns of fertilizer responsivity.
 We claim that identifying MZs characterized by specific fertilizer responsivity patterns offers valuable insights for designing treatments tailored to the specific needs of each MZ.
 
 In Fig.~\ref{fig:rcurves_fieldA}, we observe that cluster ``2" represents an MZ encompassing areas of the field characterized by low fertilizer responsivity. 
 These areas exhibit a limited reaction to changes in fertilizer rate compared to the other zones.
 In contrast, cluster ``3" represents the MZ exhibiting the highest fertilizer responsivity.
 Note that, while the curves in cluster ``4" share a similar shape with those in cluster ``3", the latter depicts a slightly steeper behavior, achieving higher yield values with lower fertilizer rates.
 Similarly, the curves depicted in Fig.~\ref{fig:rcurves_fieldB} can be associated with low, medium, and high fertilizer responsivity regions.
 In our initial experiments, we found that introducing a fourth MZ would lead to the creation of a redundant cluster with curves closely resembling those in cluster ``3". 
 We observed that employing four MZs resulted in zone boundaries that were overly intricate and impractical.

 Note that the adapted CFE generation method enables us to derive local explanations for each field location, providing insights into why they were assigned to a specific MZ.
 For instance, Fig.~\ref{fig:CFE_example} shows a site of Field A that was assigned to cluster ``2". 
 The generated CFE suggests that an increase in the $VV$ variable could enhance the site's fertilizer responsivity, leading to its clustering into another MZ.
 Given that $VV$ is associated with soil moisture content, this outcome could be interpreted as if the primary factor contributing to the site's assignment to cluster ``2" is its relatively low moisture content.

\begin{figure}[t]
    \centering
    \includegraphics[width=\columnwidth]{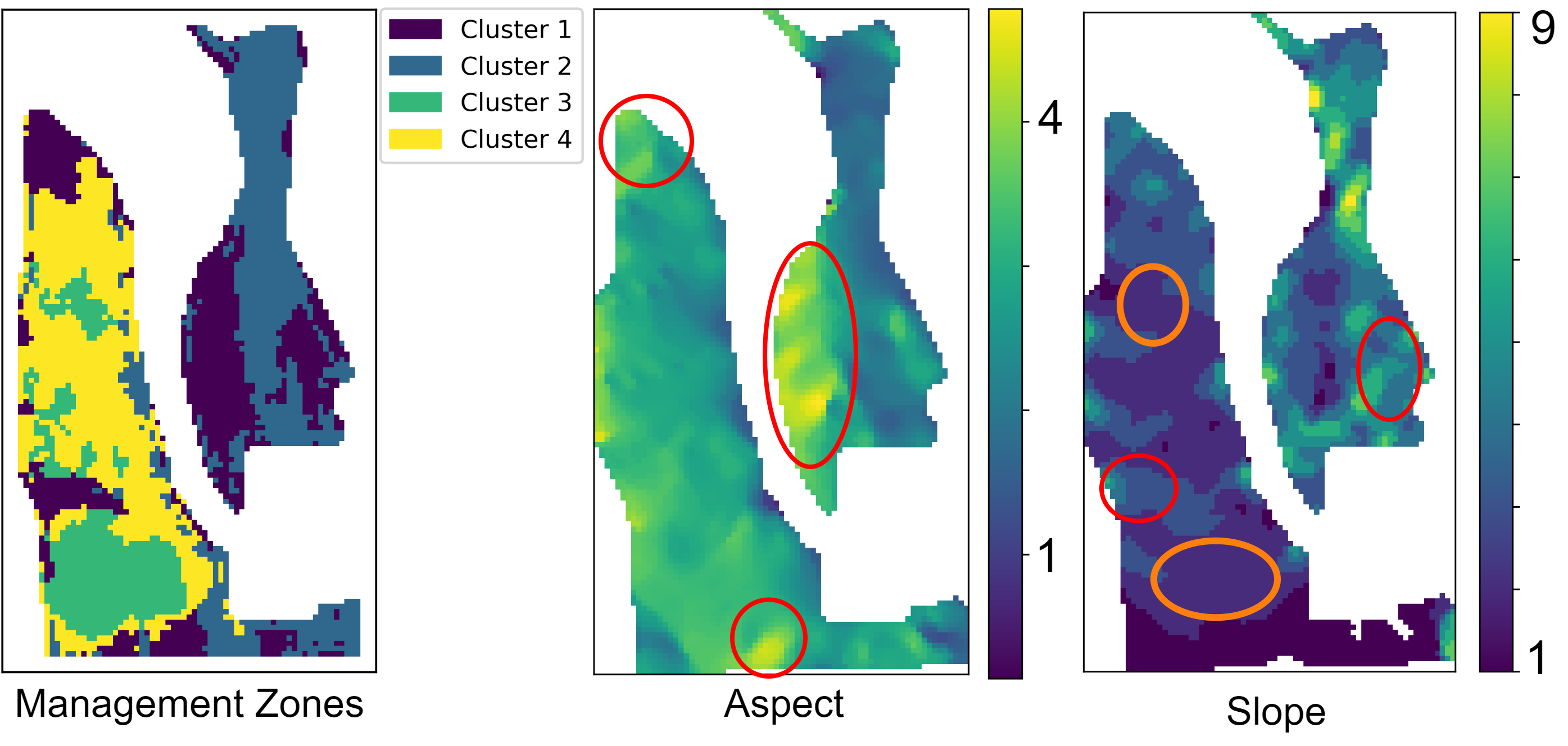}
    % \vspace{-3ex}
    \caption{Zones obtained for field A along its $A$ and $S$ maps.}
    % \vspace{-1ex}
    \label{fig:CFE_global}
\end{figure}

 Applying the CFE generation process to all field points and aggregating the results enables us to derive global explanations that characterize the overall behavior of each MZ.
 For example, from Fig.~\ref{fig:CFE_fieldA} and Table~\ref{tab:fieldA_comb}, we conclude that variables $S$, $A$, and $VV$ have the most significant influence in determining the assignment of a particular field point to MZ ``1."
 Fig.~\ref{fig:CFE_global} shows the topographic slope and the topographic aspect maps for Field A.
 From this, we observe that sites clustered in MZ ``1" coincide with areas with high aspect values and medium to high slope values, as indicated by the red ellipses.
 Note that areas with higher slope values are more prone to fertilizer runoff, which affects their fertilizer responsivity.
 In addition, high aspect values correspond to a northeast slope orientation (in the Northern Hemisphere), implying limited sunlight and increased snow retention in comparison to other regions of the field.
 From Fig.~\ref{fig:CFE_fieldA} and Table~\ref{tab:fieldA_comb}, we also conclude that the most relevant variables in determining the assignment of a particular field point to MZ ``3" (i.e., the highest fertilizer responsivity) are $TPI$ and $S$.
 Note in Fig.~\ref{fig:CFE_global} that the sites in MZ ``3" occupy areas with low slope values, as well as most of the sites in MZ ``4".
 These findings indicate that $TPI$ plays a crucial role in distinguishing the fertilizer responsivity of relatively flat regions. 
 $TPI$ is associated with the terrain's ruggedness, and higher values correspond to an increased likelihood of runoff.
 Similar explanations are obtained for the rest of the MZs.

 On the other hand, the results from Fig.~\ref{fig:CFE_fieldB} indicate that the feature relevance values are similar for the three MZs.
 This can be attributed to the uniform elevation values observed throughout Field B, where there are no steep and abrupt regions such as in Field A.
 Therefore, similar to the case of the MZ ``3" of Field A, the most important variables for cluster membership are $S$, $TPI$, and $VV$, according to Fig.~\ref{fig:CFE_fieldB} and Table~\ref{tab:fieldB_comb}.
 It is worth pointing out that, for both fields, most of the generated CFEs required changes in individual features, as evidenced in Table~\ref{tab:fieldA_comb} and Table~\ref{tab:fieldA_comb}.
 However, in the case of Field B, there was a greater need for changes involving at least two variables compared to Field A. 
 For instance, the combination of $S$ and $VV$ was the second most effective variable combination, recurring in 11\% of the CFEs generated for MZ ``1."
 This implies that, in many cases, the low responsivity of sites in MZ ``1" can be attributed not solely to the elevated risk of fertilizer runoff caused by high slope values, but concurrently to their poor moisture content.

\section{Conclusion} \label{conclusion}

The determination of fertilizer management zones allows for the design of targeted and optimized treatment strategies. 
While existing methods consider various factors, none explicitly address the crucial aspect of fertilizer responsivity in defining these zones. 
To address this gap, we introduce a novel management zone clustering method based on neural network-generated response curves that accounts for the within-field variability of fertilizer responsivity.

Our approach leverages N-response curves generated using a specialized 2D regression convolutional neural network, providing an approximation of how crop yield responds to varying fertilizer rates. 
We characterize the shape dissimilarity of these curves and use it as a metric for clustering analysis. 
Experimental results on two winter wheat dryland fields demonstrate the effectiveness of our method in organizing field points into MZs with consistent N-response curve shapes.
Thus, each MZ exhibits a distinct fertilizer responsivity pattern.

Finally, our counterfactual explanation method improves understanding of the impact of various covariate factors on the assignment of field points to specific MZs. 
Our experiments provide explanations suited to each MZ and field, uncovering the influence of specific terrain characteristics on fertilizer responsivity.
Future work will focus on the design of statistical tests that would allow for determining the optimal number of MZs.
In the context of On-Farm Precision Experimentation (OFPE), we also plan to modify our approach to assist in creating optimal fertilizer maps for the different zones. 
Specifically, instead of using the conventional rectangular gridding, we will employ several small, homogeneous regions that align better with the local variations in the field.

\section*{Acknowledgements}
% Details hidden for double-blind review purposes.
The authors wish to thank the team members of the On-Field Precision Experiment (OFPE) project for their comments throughout the development of this work.
This research was supported by a USDA-NIFA-AFRI Food Security Program Coordinated Agricultural Project 
% , titled “Using Precision Technology in On-farm Field Trials to Enable Data-Intensive Fertilizer Management,” 
(Accession Number 2016-68004-24769), and also by the USDA-NRCS Conservation Innovation Grant from the On-farm Trials Program
% , titled “Improving the Economic and Ecological Sustainability of US Crop Production through On-Farm Precision Experimentation” 
(Award Number NR213A7500013G021).

\balance
\bibliographystyle{IEEEtran}
\bibliography{references}

\end{document}